\documentclass{amia}
\usepackage{graphicx}
\usepackage[labelfont=bf]{caption}
\usepackage[superscript,nomove]{cite}
\usepackage{color}
\usepackage[super,comma,sort&compress]{natbib}

\begin{document}

\title{The Dependence of Machine Learning on Electronic Medical Record Quality}
\author{Long V. Ho, David Ledbetter, Melissa Aczon, Ph.D., Randall Wetzel, M.D.}

\institutes{
	The Laura P. and Leland K. Whittier Virtual Pediatric Intensive Care Unit \\
    Children's Hospital Los Angeles, Los Angeles, CA\\
}

\maketitle

\noindent{\bf Abstract}



\textit{There is growing interest in applying machine learning methods to Electronic Medical Records (EMR). Across different institutions, however, EMR quality can vary widely. This work investigated the impact of this disparity on the performance of three advanced machine learning algorithms: logistic regression, multilayer perceptron, and recurrent neural network. The EMR disparity was emulated using different permutations of the EMR collected at Children's Hospital Los Angeles (CHLA) Pediatric Intensive Care Unit (PICU) and Cardiothoracic Intensive Care Unit (CTICU). The algorithms were trained using patients from the PICU to predict in-ICU mortality for patients in a held out set of PICU and CTICU patients. The disparate patient populations between the PICU and CTICU provide an estimate of generalization errors across different ICUs. We quantified and evaluated the generalization of these algorithms on varying EMR size, input types, and fidelity of data.}

\section*{Introduction}

Electronic Medical Records (EMR) are currently adopted by approximately {84\%} of hospitals in the United States\cite{henry2016adoption}. With the recent success of machine learning, there are numerous research efforts to extract medically relevant and actionable information from troves of EMR using machine learning algorithms\cite{ghassemi2015state}. However, due to varying data entry protocols and bespoke implementations of EMR systems, data quality can vary widely across institutions\cite{cios2002uniqueness}. This work investigated the dependence of advanced machine learning algorithms on the varying EMR quality.


EMR data discrepancies were emulated using permutations of EMR collected from Children's Hospital Los Angeles (CHLA) Pediatric Intensive Care Unit (PICU) and Cardiothoracic Intensive Care Unit (CTICU). The effects of the permutations were analyzed on three machine learning algorithms: logistic regression, multilayer perceptron, and recurrent neural network. Permutations studied included varying the number of patient encounters available for training; which variables were included and excluded as model inputs; and varying the fidelity of drug information.

The task used to evaluate these algorithms was prediction of in-ICU mortality: whether the patient survived their ICU encounter. Mortality was chosen because it provides an objective measure of a model's ability to extract information from EMR\cite{knaus1985apache}. Predictions were generated for each algorithm after 12 hours of observation and used to create Area Under the Receiver Operating Curves (AUROC).
\section*{EMR Source} \label{data_source}
This study used anonymized EMR collected in the Pediatric Intensive Care Unit (PICU) and Cardiothoracic Intensive Care Unit (CTICU) at Children's Hospital Los Angeles (CHLA) between 2002 and 2016. The database contained 21,881 ICU encounters (defined as the time between a patient's admission to discharge from the ICU). Of these ICU encounters, 16,706 were from the PICU (12,093 patients with {4.85\%} mortality) and 5,175 were from the CTICU (3,088 patients with {3.32\%} mortality). Each encounter contained irregularly charted measurements of the patient's physiology (e.g. heart rate, respiratory rate), laboratory test results (e.g. glucose, creatinine), and treatments (e.g. intubation, epinephrine) administered throughout their ICU stay. Additionally, each encounter was accompanied by static information describing demographics (e.g. age, sex, and race), diagnosis, and disposition at the end of the ICU encounter (whether they survived). 

\section*{Pre-Processing EMR for Machine Learning} \label{data_preproc}
The collected EMR was a list of observations and treatments (in long format) charted by the clinical team, timestamped and uniquely identified to each patient. The following sub-sections describe pre-processing techniques that restructured the EMR for machine learning purposes.\newline

\textit{Data Curation:} 
In collaboration with physicians, the dataset was curated to remove erroneous observations and aggregate variables of similar groups. For example, minimum and maximum values were defined for each observation such that erroneous measurements which were incompatible with life (e.g. heart rate of 1200 beats per minute) were corrected. Additionally, different observations of similar variables such as invasive and non-invasive blood pressures were combined into a single variable when medically appropriate\cite{imholz1990non}.

\begin{figure}[htbp!]
\centering
\includegraphics[height=1.3in, width=6.5in]{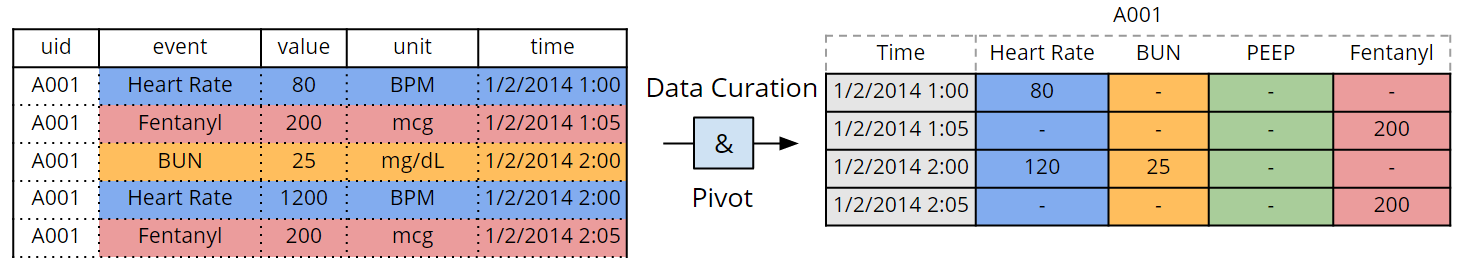}
\caption{Illustration of pivot operation (and curation) applied to long format EMR to create a patient-matrix. Note that for a particular patient, a column of the pivoted data may have a blank entry in each row as in PEEP above.\newline}
\label{fig:data_pivot}  
\end{figure}

\textit{Data Pivot:} 
To accommodate the structure required for machine learning algorithms, the long format EMR was pivoted to wide format. In other words, the EMR was reshaped such that the timestamps were the rows, observations were the columns, and values populated the cells. The result was a sparse, irregularly sampled matrix for each patient, henceforth referred to as a patient-matrix. Figure \ref{fig:data_pivot} illustrates an example of this process.

\textit{Data Standardization:} 
Physiological observations (labs and vitals) were converted to z-scores using the means and standard deviations computed from the \textit{training set} (described later). Treatments administered to the patients such as drugs and interventions were normalized to values between [0, 1] using upper limits defined in collaboration with clinicians.

\begin{figure}[htbp!]
\centering
\includegraphics[height=1.2in, width=6.5in]{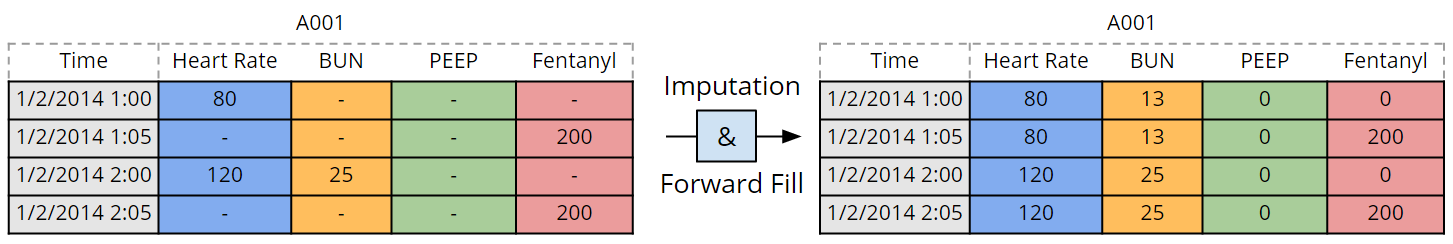}
\caption{Illustration of imputing a patient-matrix. Data standardization is not applied here to show realistic values.\newline}
\label{fig:data_imputation}
\end{figure}

\textit{Data Imputation:} 
Physiologic observations were forward-filled following an initial measurement until the next available measurement. This choice was based on the clinical insight that measurements are more frequently taken during times of hemodynamic instability and less frequently when the patient appears stable. If a physiologic variable had no recorded measurements for an entire encounter, that variable was set to zero. Because of the prior standardization of physiologic variables, this imputation was equivalent to imputing the population mean derived from the training set. For drugs and interventions, missing values were imputed with zero. Since these variables were normalized to [0, 1], a zero entry indicated the absence of treatment. Figure \ref{fig:data_imputation} depicts the imputation process.

\textit{Train, Validation, Test Partitioning:} 
The data were partitioned into train, validation, and test sets. The training set consisted of {50\%} percent of the PICU patients  ({8,404} encounters, {4.66\%} mortality), selected at random. Half of the remaining PICU patients ({4,122} encounters, {4.97\%} mortality) constituted the validation set used to tune the algorithm's hyper-parameters. Model performance was evaluated on two test sets: the last 25\% of the patients from the PICU ({4,176} encounters, {5.12\%} mortality); and all of the patients in the CTICU ({5,175} encounters, {3.32\%} mortality). These test sets were not used during any model training and development. Since the CTICU specializes in congenital heart defects and the PICU treats general critical cases, the CTICU was used as a surrogate to represent an institution with disparate patient population and treatments. Scores on the PICU test set show the algorithms' generalization to similar ICUs while scores on the CTICU test set estimate performance across different ICUs.
\section*{Data Permutations}
Permutations of CHLA EMR were generated to emulate the data disparity found across different EMR systems. The following subsections describe three types of permutations: training set fraction (varying data size), input types (varying input variables), and drug encoding (varying data fidelity). 

\textit{Training Set Fraction:} 
A common concern with applying machine learning techniques such as neural networks is the amount of data required for training the models to perform well \cite{walczak2001empirical,flexer1996statistical}. This question is particularly relevant 
because the number of patient records can vary widely across hospitals. Two primary factors contribute to the disparity: 1) the size of the institution (number of critical care beds available) and; 2) the institution's date of EMR adoption. To assess the dependence of algorithm performance on training data size, the algorithms were trained on successively decreasing sizes -- 100, 90, 75, 50, 25, 10\% -- of the full training dataset described earlier. The resulting model from each of these trainings was then evaluated on the same complete PICU and CTICU test sets. The validation set remained fixed. 

\begin{figure}[!htbp]
\centering
\includegraphics[height=1.7in, width=4.5in]{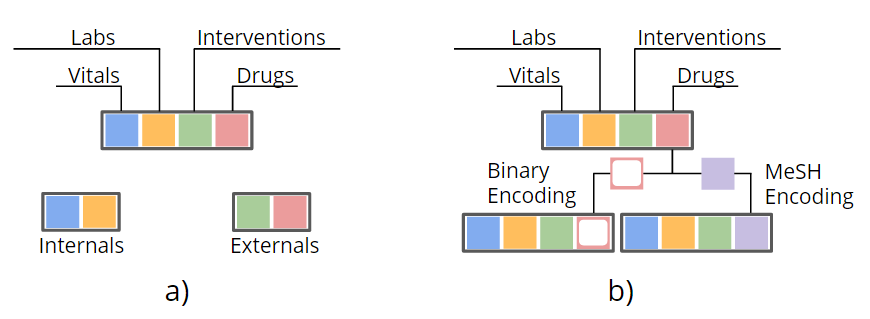}
\caption{The complete patient-matrix contains information about the patient's internals (vitals, labs) and externals (drug administrations and interventions). a) Illustrates parsing of internals and externals from the complete patient-matrix. b) Shows the encoding of the drug subset in patient-matrix to the two permutation schemes: binary drug indicator and MeSH.}
\label{fig:data_permutations}
\end{figure}

\textit{Input Types:} 
The medical community has had many discussions on the validity of incorporating treatments and interventions into patient acuity scores \cite{pollack2016severity, pollack2016pediatric}. Since treatments typically are not applied randomly, their application is informative of patient state.
However, the `insight' or value of such information is complicated by non-standard practices across clinicians when administering treatments 
\cite{anthony2013big}. 
Conversely, the same physiologic measurement can have different interpretations that depend on the treatments being received. For instance a heart rate may appear normal in isolation, but 
the same heart rate in the presence of vasopressors may indicate a higher risk of mortality. This permutation study parses the dependence of the algorithms on internal and external inputs. Internal refers to the observations describing the patient's status (labs and vitals) and external refers to the inputs that describe the treatments given to the patient (drugs and interventions). Figure \ref{fig:data_permutations}a illustrates the splitting of the inputs from baseline (combined) to internal and external inputs. 

\textit{Drug Encoding:}
The fidelity of the captured information may vary from institution to institution. For instance, data may be limited and contain only an indication of whether or not a particular drug was administered, but not its exact quantity. Moreover, treatment protocols vary from institution to institution. For example, one institution may preferentially provide albumin infusions while another may favor saline for fluid therapy \cite{finfer2004comparison}. This permutation experiment emulates the variance in data fidelity using two drug encoding schemes: 1) decimating drug values to binary indicators; and 2) indexing the drugs based on biomedical ontologies by encoding them to the National Library of Medicine's Medical Subject Headings (MeSH). Figure \ref{fig:data_permutations}b illustrates the baseline input's encoding to these two schemes.\newline

\section*{Methods}
\begin{figure}[!htbp]
\centering
\includegraphics[height=3.45in, width=4.45in]{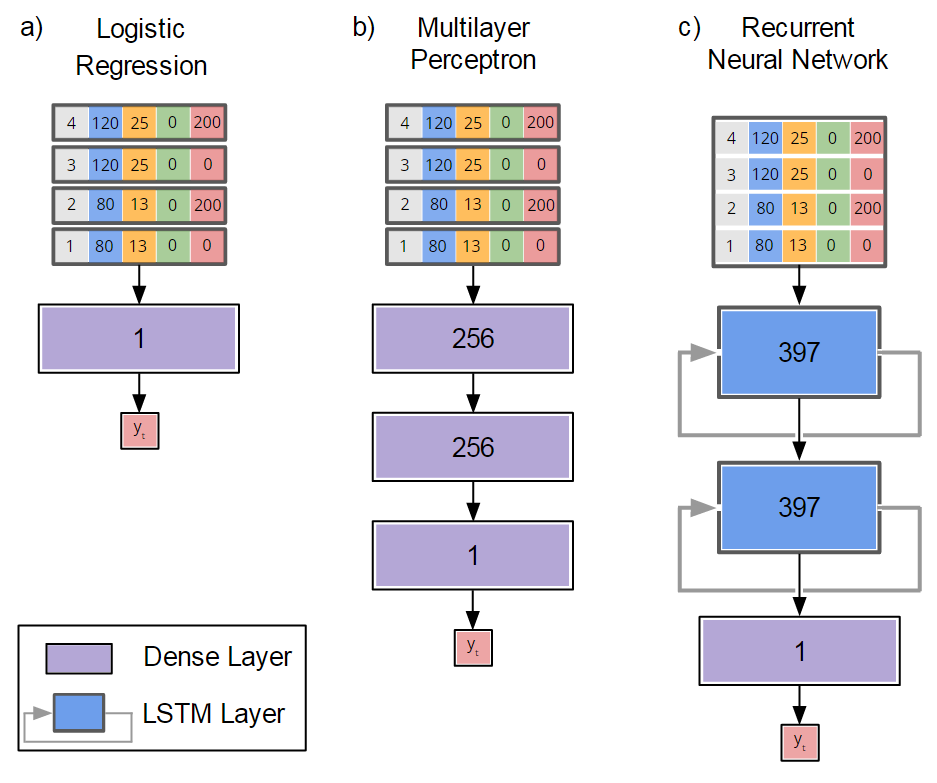}
\caption{Model architectures for a) logistic regression, b) multilayer perceptron, and c) recurrent neural network. Each model is composed of layers that operate on the patient's input sequences (multi-colored vectors) to predict the patient's in-ICU mortality. The numbers in each layer (purple dense layer and blue LSTM layer) are the number of hidden units of that layer. \newline} 
\label{fig:model_architectures}
\end{figure}

Three representation learning algorithms were assessed: logistic regression (LR), multilayer perceptron (MLP), and recurrent neural network (RNN). The LR and MLP have been successful in a variety of medical applications including detection of prostate cancer, detection of Acute Respiratory Distress Syndrome (ARDS), and estimation of childhood asthma risk \cite{langer2009prostate, force2012acute, moffatt2007genetic}. RNNs have recently shown success in medical tasks such as predicting critical decompensation of patients on the floor \cite{shah20162}, early detection of heart failure \cite{choi2016using}, and predicting ICU mortality \cite{aczon2017dynamic}. The following sub-sections describe more specifically the variant of each algorithm used. Figure \ref{fig:model_architectures} shows the architecture of the three models.

\textit{Logistic Regression:}
The logistic regression is one of the simplest machine learning models. 
It assumes that the input ($x$) and output data ($y$) are related by the mapping
$y=\sigma \left(W^{\top}x \right)$, where $\sigma(x)=1/\left(1+e^{-x}\right)$. The matrix of weights, $W$, is learned from the data. For our task, the input to the LR is the pre-processed EMR derived patient-matrix, and the output is the patient's mortality response, i.e. disposition (alive or not) at the end of ICU encounter. Although the input is a patient-matrix, the LR operates on individual \textit{time-slices} (a vector) of the patient matrices.

\textit{Multilayer Perceptron:}
The MLP, also called a feed forward neural network, is a classic deep learning method 
which models the mapping from inputs to output with $y = f^{*}(x)$, where $f^{*}$ is a series of different functions called layers. The construction is 
loosely inspired by biological neural networks.
Each node (hidden unit) in a layer is associated with a weight, $w_{i, j}$, that connects to every node in the subsequent layer, analogous to the communication between neurons. The MLP architecture used here consisted of a 2-layer MLP, each with hidden units of 256. Note that the MLP also operates on time-slices of the patient matrix, and that the LR described above is equivalent to a 0-layer MLP when trained similarly using mini-batch stochastic gradient descent. 

\textit{Recurrent Neural Network:} 
The RNN refers to the family of neural networks specialized for processing sequential data \cite{funahashi1993approximation,goodfellowbook}. The RNN can be regarded an extension of the MLP that contains connections which feed each layer's outputs back into itself. This recurrent structure makes RNNs ideal for processing EMR because the hidden states and feedback loop can elegantly integrate and incorporate past information about the patient with newly acquired measurements \cite{aczon2017dynamic}. This characteristic also makes it a fitting choice for modeling the dynamic disease progression encountered in a critical care setting. The particular RNN variant used is the Long Short Term Memory (LSTM)\cite{hochreiter1997long}. The model is composed of a 2-layer LSTM, each with 397 hidden units (equal to the number of variables in the input patient-matrices). The output layer is a dense layer applied to every time-step of the output from the previous LSTM layer. Compared to the LR and MLP, the RNN operates on the entire patient-matrix.

The above three models were trained using mini-batch sizes of 128 and optimized using RMSprop \cite{tieleman2012lecture}, a gradient descent optimizer that utilizes an adaptive learning rate, to minimize the binary cross-entropy between the model's predicted in-ICU probability and the patient's true mortality response. Each model's layers were initialized using Glorot uniform, a method that samples weights from a uniform distribution with the variance scaled based on the number of parameters of the preceding and subsequent layers. After every epoch of training (an epoch is defined as one cycle through the training set), the models predicted the in-ICU mortality for patients in the validation set and the binary cross-entropy was computed. If the binary cross-entropy did not decrease after 15 epochs, the model's learning rates were decimated by 5. After 2 reductions, training is stopped and the weights associated with the best validation performance are used to predict the mortality of the patients in the two test sets. During training, dropout of {20\%} was applied to the input sequences (across time) as a data augmentation parameter.


\section*{Results} \label{results}
Model training for each data permutation study was iterated five times, with each iteration corresponding to a different initialization point (for the model weights) and different validation stopping point. The resulting model from each iteration was then applied to the PICU and CTICU test sets, and the Area Under the Receiver Operating Curve (AUROC) was generated from the model's in-ICU mortality prediction after 12 hours of observation. The numbers reported in Tables \ref{mortAUCdatadeg}--\ref{mortAUCdrugencoding} are the mean and standard deviations of the AUROCs from these iterations.
\textit{Baseline} (BL) refers to the processed-EMR containing all available training encounters and input variables. Baseline is the same across the different EMR permutation studies and occupies the first row of Tables \ref{mortAUCdatadeg}--\ref{mortAUCdrugencoding} while being named accordingly ({100\%}, Combined, NoEncoding).\newline

\begin{table*}[!htbp]
\centering
\caption{In-ICU mortality AUC on the test sets as a function of training set fraction used. The numbers in parenthesis are the resulting number of encounters used in each training subset fraction.}
\label{mortAUCdatadeg}
\resizebox{\columnwidth}{!}{%
\begin{tabular}{l|ccc|ccc}
\hline
 \textbf{Training}            & \multicolumn{3}{c|}{\textbf{PICU Test Set}}                              & \multicolumn{3}{c}{\textbf{CTICU Test Set}}                              \\
 \textbf{Set Fraction}       & \textbf{LR}      & \textbf{MLP}              & \textbf{RNN}             & \textbf{LR}      & \textbf{MLP}              & \textbf{RNN}             \\ \hline
\textbf{100\% (BL)}    & 0.907 +/- 0.001 & 0.909 +/- 0.001          & 0.921 +/- 0.003 & 0.803 +/- 0.002 & 0.808 +/- 0.001 & 0.801 +/- 0.008          \\
\textbf{75\% (6306)}     & 0.899 +/- 0.001 & 0.904 +/- 0.001          & 0.914 +/- 0.001 & 0.808 +/- 0.004 & 0.813 +/- 0.001 & 0.809 +/- 0.011          \\
\textbf{50\% (4204)}     & 0.894 +/- 0.002 & 0.902 +/- 0.001          & 0.905 +/- 0.002 & 0.795 +/- 0.002 & 0.807 +/- 0.002 & 0.791 +/- 0.003          \\
\textbf{25\% (2102)}     & 0.890 +/- 0.002 & 0.896 +/- 0.001          & 0.904 +/- 0.003 & 0.790 +/- 0.002 & 0.804 +/- 0.002 & 0.793 +/- 0.007          \\
\textbf{10\% (840)}     & 0.867 +/- 0.002 & 0.891 +/- 0.001 & 0.881 +/- 0.003          & 0.773 +/- 0.002 & 0.775 +/- 0.003          & 0.783 +/- 0.011 \\
\end{tabular}
}
\end{table*}

\begin{table*}[!htbp]
\centering
\caption{In-ICU mortality AUC on the test sets as a function of input types used.}
\label{mortAUCmeastype}
\resizebox{\columnwidth}{!}{%
\begin{tabular}{l|ccc|ccc}
\hline
  \textbf{Input}                 & \multicolumn{3}{c|}{\textbf{PICU Test Set}}                     & \multicolumn{3}{c}{\textbf{CTICU Test Set}}                              \\
   \textbf{Types}                 & \textbf{LR}      & \textbf{MLP}     & \textbf{RNN}             & \textbf{LR}              & \textbf{MLP}              & \textbf{RNN}    \\ \hline
\textbf{Combined (BL)}  & 0.907 +/- 0.001 & 0.909 +/- 0.001 & 0.921 +/- 0.003 & 0.803 +/- 0.002         & 0.808 +/- 0.001 & 0.801 +/- 0.008 \\
\textbf{Internals} & 0.899 +/- 0.001 & 0.903 +/- 0.001 & 0.917 +/- 0.002 & 0.801 +/- 0.002          & 0.808 +/- 0.002 & 0.793 +/- 0.005 \\
\textbf{Externals} & 0.841 +/- 0.001 & 0.833 +/- 0.002 & 0.841 +/- 0.002 & 0.727 +/- 0.003 & 0.721 +/- 0.001          & 0.679 +/- 0.005
\end{tabular}
}
\end{table*}

\begin{table*}[!htbp]
\centering
\caption{In-ICU mortality AUC on the test sets as a function of the drug encoding used.}
\label{mortAUCdrugencoding}
\resizebox{\columnwidth}{!}{%
\begin{tabular}{l|ccc|ccc}
\hline
 \textbf{Drug}                 & \multicolumn{3}{c|}{\textbf{PICU Test Set}}                     & \multicolumn{3}{c}{\textbf{CTICU Test Set}}                     \\
 \textbf{Encoding}                 & \textbf{LR}      & \textbf{MLP}     & \textbf{RNN}             & \textbf{LR}      & \textbf{MLP}              & \textbf{RNN}    \\ \hline
\textbf{None (BL)} & 0.907 +/- 0.001 & 0.909 +/- 0.001 & 0.921 +/- 0.003 & 0.803 +/- 0.002 & 0.808 +/- 0.001 & 0.801 +/- 0.008 \\
\textbf{Binary}   & 0.909 +/- 0.001 & 0.912 +/- 0.001 & 0.922 +/- 0.002 & 0.803 +/- 0.004 & 0.807 +/- 0.001 & 0.794 +/- 0.009 \\
\textbf{MeSH}     & 0.906 +/- 0.001 & 0.910 +/- 0.001 & 0.922 +/- 0.001 & 0.782 +/- 0.001 & 0.785 +/- 0.002 & 0.752 +/- 0.007
\end{tabular}
}
\end{table*}

\begin{figure}[!htbp]
\centering
\includegraphics[height=3.0in, width=6.25in]{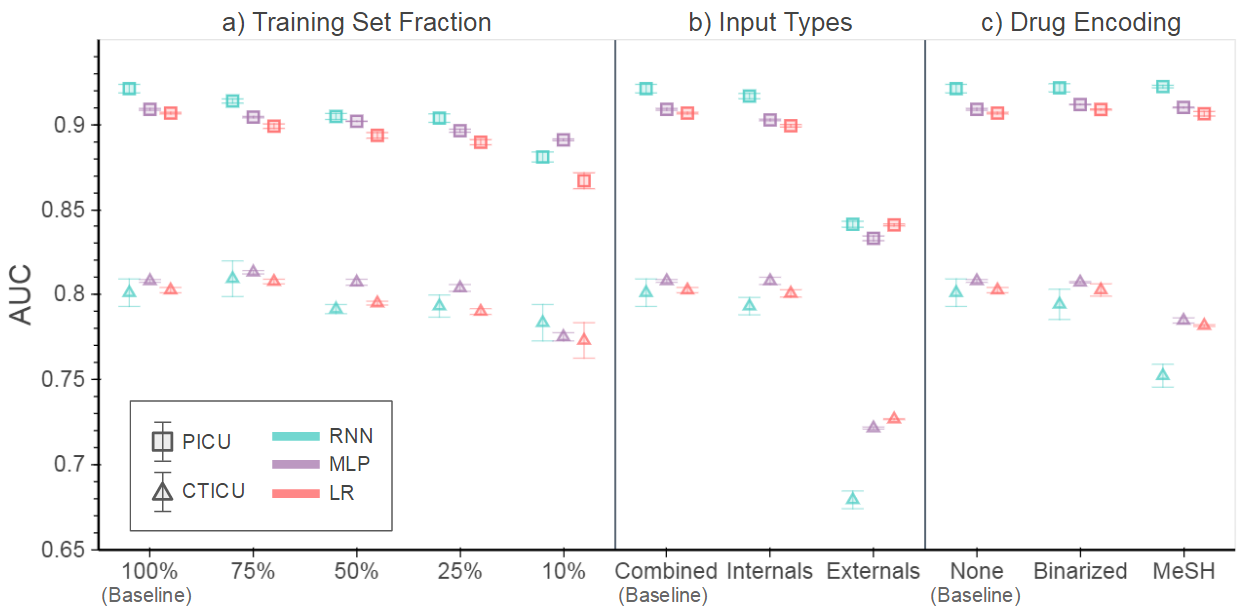}
\caption{In-ICU mortality AUC on the test sets for the three EMR permutations: a) training set fraction, b) input types, and c) drug encoding. Squares denote performance on the PICU test set and triangles correspond to the CTICU test set. The vertical lines are the computed standard deviations. The colors distinguish the models: cyan (RNN), purple (MLP), and red (LR).\newline}
\label{fig:resultsv1}
\end{figure}

Table \ref{mortAUCdatadeg} shows the performance of each algorithm as a function of the number of patients in the training set. Table \ref{mortAUCmeastype} shows the performance when limiting the model inputs to either only internal (physiologic observations) or only external (treatments) variables. Table \ref{mortAUCdrugencoding} shows the performance as a function of drug encoding strategy. Figure \ref{fig:resultsv1} offers a visual summary of Tables \ref{mortAUCdatadeg}--\ref{mortAUCdrugencoding}.  As a point of reference, AUROCs were also computed from the predictions of  two standard pediatric severity of illness scoring systems, PIM2\cite{slater2003pim2} and PRISM3-12\cite{pollack1996prism}.  PIM2's AUROCs on the PICU and CTICU test sets were {0.868} and {0.774}, while those of PRISM3-12 were {0.856} and {0.723}, respectively.

\section*{Discussion} \label{discussion}

A large performance disparity was observed between the PICU and CTICU test sets across all data permutations and algorithms. The performance on CTICU data was approximately 10\% lower than on the PICU data. This is not surprising since the models were trained on PICU encounters only, and the CTICU patient population and treatments can be very different from those in the PICU.  On the PICU test set, the RNN models substantially and consistently outperformed their LR and MLP counterparts across the different permutation experiments, with the lone exception coming from the experiment using only 10\% (840 encounters) of the training data for model development.  On the CTICU test set, the MLP models consistently outperformed their RNN and LR counterparts; the lone exception was again observed in the 10\% of training data experiment, where RNN outperformed MLP. The results indicate that the MLP learned more generalizable patterns than the RNN, despite the latter's more advanced sequence-processing capabilities. The MLP proved more robust, relative to the RNN, in transferring across disparate ICUs.  The RNN's large number of parameters ($\sim$15 times more trainable parameters) may have led to overfitting to the PICU data.


Table \ref{mortAUCdatadeg} and Figure \ref{fig:resultsv1}a illustrate reduction in each model's performance on both PICU and CTICU as the amount of training data was reduced. When trained using only 840 encounters (training fraction of {10\%}), the models perform reasonably well on the {$\sim$4000} encounters in the PICU and CTICU test sets, with the weakest performing model (LR) scoring 0.867 and 0.773 on the PICU and CTICU, respectively. 

Table \ref{mortAUCmeastype} and Figure \ref{fig:resultsv1}b show little performance degradation  when external information was removed from the baseline inputs. This indicates that internal information contributed almost entirely to the each model's prediction of the patient's severity of illness. Although externals alone performed significantly lower than internals and baseline, the performance increase from internals to baseline indicates that the models extracted additional information from the external variables.


The results in Table \ref{mortAUCdrugencoding} and Figure \ref{fig:resultsv1}c show that model performance, regardless of algorithm, remained about the same when changing drug information from real-valued to binary indicators (absent or present). 
This similar performance may be 
due to standard quantities and protocols frequently being followed during drug administration \cite{dellinger2013surviving}.
Consequently, 
knowledge of a clinical decision to apply a particular drug, regardless of its dosage, may be sufficient to understand the patient's physiologic response to that drug.

Table \ref{mortAUCdrugencoding} and Figure \ref{fig:resultsv1}c further show that on the PICU set, the models using baseline drug encoding performed similarly to models using 
MeSH-encoded drug information.
On the CTICU set, however, the degradation of RNN performance from its baseline model to the one using MeSH was 5\%.  The RNN's significantly worse degradation on the CTICU (compared to slight degradation on the PICU test set) was also observed in the baseline-to-externals experiment (Table 2 and Figure 5b).  
This pattern may be due to the RNN overfitting the information gleaned from external variables. The LR and MLP models  
process a time-slice in isolation of the other time-slices in a patient-matrix, but the RNN builds a history of the applied treatments.  This sequential processing of data likely made the RNN learn to weigh external information more. Since the CTICU contained a more disparate distribution in treatments, over-utilizing the information learned from the PICU may have improved its performance on the PICU but hindered its generalization to the CTICU.

\section*{Conclusions}
This work measured the effects of varying EMR quality on the performance of several advanced machine learning algorithms: logistic regression, multilayer perceptron, and recurrent neural network. Three sets of permutations of CHLA EMR were generated to emulate the EMR data disparity across institutions. The first measures the performance of the models as a function of training data size. The second measures the dependence of the models on varying input types. 
The third measures the effects of varying fidelity in drug data.
The algorithms were analyzed by measuring their performance in predicting in-ICU mortality in two different ICUs. 

%
Performance of all three models degraded with decreasing size of the training set.  Even when trained on 10\% of the available encounters (840), all models still performed comparably with two standard pediatric severity of illness scores, PIM2 and PRISM3-12. Additionally, the MLP generalized better than the RNN when they were trained on PICU but tested on CTICU data.

Future work includes measuring the effects of the machine learning models on other varying types of EMR. For example, measuring the difference in performance due to varying \textit{temporal} fidelity by incorporating waveform measurements from bedside monitors. Moreover, the analysis can be further strengthened by analyzing the performance on other clinically relevant tasks such as predicting ICU readmission.

\section*{Acknowledgements}
We are grateful to the Laura P. and Leland K. Whittier Foundation for funding this work. We are also thankful to Sareen Shah, M.D. for providing significant medical domain expertise; Alysia Flynn for providing assistance in aggregating data; and Jon Williams and Alec Gunny for technical discussion and comments.

\makeatletter
\renewcommand{\@biblabel}[1]{\hfill #1.}
\makeatother

\bibliographystyle{vancouver}
\bibliography{bibliography}
\end{document}